\documentclass[journal]{IEEEtran}

\usepackage{lineno,hyperref}
\usepackage[usenames, dvipsnames]{xcolor}
\usepackage{soul}
\usepackage{amsmath}
\usepackage{float}
\usepackage{subcaption}
\usepackage{adjustbox}
\modulolinenumbers[5]
\usepackage{tabularx,booktabs}
\usepackage{todonotes}
\usepackage{svg}
\usepackage{pdfpages}

\usepackage{caption}
\usepackage{subcaption}

\definecolor{sand}{RGB}{255,255,187}
\definecolor{shale}{RGB}{142,142,139}

\usepackage{amsfonts}
\usepackage{ulem}

\begin{document}

\title{Non-contrastive representation learning for intervals from well logs}

\author{Alexander Marusov,
        Alexey Zaytsev

\thanks{A. Marusov, A. Zaytsev are with Skolkovo Institute of Science and Technology (Skoltech), 121205. This work was supported by Ministry of Science and Higher Education grant No. 075-10-2021-068.}}


\maketitle

\begin{abstract}
The representation learning problem in the oil \& gas industry aims to construct a model that provides a representation based on logging data for a well interval. Previous attempts are mainly supervised and focus on similarity task, which estimates closeness between intervals. We desire to build informative representations without using supervised (labelled) data. 

One of the possible approaches is self-supervised learning (SSL). In contrast to the supervised paradigm, this one requires little or no labels for the data. Nowadays, most SSL approaches are either contrastive or non-contrastive. Contrastive methods make representations of similar (\textit{positive}) objects closer and distancing different (\textit{negative}) ones. Due to possible wrong marking of positive and negative pairs, these methods can provide an inferior performance. Non-contrastive methods don't rely on such labelling and are widespread in computer vision. They learn using only pairs of similar objects that are easier to identify in logging data. 

We are the first to introduce non-contrastive SSL for well-logging data. In particular, we exploit Bootstrap Your Own Latent (BYOL) and Barlow Twins methods that avoid using negative pairs and focus only on matching positive pairs. The crucial part of these methods is an augmentation strategy. Our augmentation strategies and adaption of BYOL and Barlow Twins together allow us to achieve superior quality on clusterization and mostly the best performance on different classification tasks. Our results prove the usefulness of the proposed non-contrastive self-supervised approaches for representation learning and interval similarity in particular.

\end{abstract}


\IEEEpeerreviewmaketitle

\section{Introduction}
Let us start with the interwell correlation~\cite{Romanenkova2022} . It is one of the crucial problems in the oil \& gas industry. We aim to understand how particular intervals in wells are similar to each other. An effective model of interwell correlation lets geologists plan oil production more effectively and detect additional risks of abnormal wells in advance. Moreover, hydrocarbon reserves estimation and building a geological field model need correct interwell correlation. However, changing sedimentation conditions across the basin makes the process of interwell correlation rather tricky. Also, the manual interwell comparison is complex and tedious work~\cite{Verma2014}. Besides, such analysis is subjective because geologists can interpret well-logging data differently.

Recently, more and more machine learning algorithms have been applied in the oil \& gas domain~\cite{Santos2021}, e.g. in rock type detection~\cite{Romanenkova2019},~\cite{Kadkhodaie-Ilkhchi2010}. Also, such kind of approach is used for the similarity learning task. The authors of~\cite{Akkurt2018} propose using Overlap and Jaccard distances between well's representations received from the SVM (Support Vector Machine) algorithm. The paper~\cite{Ali2021} is based on the same idea but has another predictor. The next paper~\cite{Rogulina2022} follows classical supervised learning principles. The feature space is the aggregated statistics of time intervals, and the target shows whether well's intervals are similar or not. 

Moving on to deep learning algorithms, the work~\cite{Cheng2020} suggests using LSTM (Long Short-Term Memory) neural network to estimate the correlation between production and injection wells. To exploit the power of CNN (Convolutional Neural Network) with working with images, the authors of the paper~\cite{Brazell2019} propose to use deep CNN for working with well's images. The supervised paradigm of the approach still requires labels generated by experts.
Authors of~\cite{Romanenkova2022} consider different modern neural network architectures (Siamese, Triplet).
They need no labels and work in a self-supervised regime.
However, their models provide imperfect scores, suggesting that better approaches should exist to capture more complex correlations and provide more universal representations, which will be helpful for the interwell correlation task. 

Obtaining a universal description of an object is a challenging but essential task, which becomes even more critical since, today, most of the actual logging data is unlabeled~\cite{Romanenkova2022}. These representations may be used in various tasks. One is \textit{clustering}, a type of \textit{similarity learning} problem. Similar intervals should have close representations and different intervals – distant representations. Hence, the idea is to cluster constructed representations (embeddings). 
Also, we can use them to solve \textit{downstream tasks} like dominant geological formation identification over a well interval. To get representations, we can use SSL. SSL is a type of unsupervised learning where the pseudo labels are taken from the raw data. For example, in~\cite{Mikolov2013}, authors hide the word in a sentence and predict it by its context. 
In our case, the dataset is unlabeled, and expert labelling is time-consuming and expensive. Instead, we create pseudo labels for well similarity based on simple rules. This approach allows us to use self-supervised methods, which can be divided into two groups: generative or discriminative~\cite{Egorov2022}.
Discriminative methods consist of contrastive and non-contrastive methods. Contrastive methods use positive and negative pairs for training. One of the most famous frameworks in computer vision is SimCLR~\cite{Chen2020}. It combines an additional projection head on top of the encoder output and contrastive loss. In the time series domain, authors of~\cite{Yue2022} propose to use the TS2Vec framework, which provides a universal representation for any time series data. However, contrastive methods have limitations. As mentioned above, one of the main problems of these approaches is a complex labelling process. Also, these methods often need to compare each example with many others, which is time-consuming~\cite{Grill2020}.


The recent works from the image domain use a non-contrastive paradigm championed by~\cite{Zbontar2021}. Non-contrastive methods use only positive pairs for training, usually obtained from a single object's augmentation (views), as we have no labels. For example, BYOL~\cite{Grill2020} and  Barlow Twins~\cite{Zbontar2021} methods are straightforward to use if we know how to define pairs of similar objects. 



The methods from these works are based on learning to compare initial and augmented versions of the data, so selecting an augmentation strategy is the crucial point in model performance.
There is a lot of research on image augmentations~\cite{Yarats2020}. However, time series data augmentations coupled with deep learning received limited attention. The paper~\cite{Iwana2021} compares different augmentation strategies on a time series classification task. The authors of~\cite{Wen2020} evaluate augmentations on several time series tasks like classification, anomaly detection and forecasting.

This paper considers state-of-the-art non-contrastive approaches like BYOL and Barlow Twins, which do not require negative pairs and perform better than contrastive algorithms. Using these methods, we train a deep learning model that considers the structure and properties of well-logging data and builds an informative representation of it. We use self-supervised methods because only unlabelled data is available for our task.



The main contributions of the work are the following:
\begin{enumerate}
    \item Non-contrastive state-of-the-art architectures, BYOL and Barlow Twins, were adapted to well-logging data. According to our knowledge, we are the first to implement those approaches to the current data type.
    \item We make the approach work by proposing an augmentation strategy that outperforms others and is simple to use.
    \item To evaluate the quality of obtained embeddings from BYOL and Barlow Twins, we construct a test bed that shows the strengths and weaknesses of obtained embeddings for a wide range of downstream problems. This testbed can be reused to compare other approaches and uses only open data.
    \item Our approaches complement each other. Representations from \textbf{BYOL} share first place with contrastive methods on \textit{clusterization task}, and \textbf{Barlow Twins} show superior performance in downstream \textit{classification problems} of identifying well's interval type.
\end{enumerate}


\section{Methods}

We start this section with the problem statement.
Then we discuss the non-contrastive paradigm and its vital component - possible augmentations. After that, we move on to Barlow Twins, and BYOL approaches. Also, we include a separate subsection devoted to technical details to make the presented results reproducible.

\subsection{Problem statement}

The general purpose of representation learning is to create an informative description of input data. Consequently, we aim to construct a model which has the following description:
\begin{itemize}
    \item{Input:} Multivariate logging data from well's interval, which can have missing values
    \item{Output:} A vector representation of the interval suitable for the solution of a wide range of problems
\end{itemize}
We expect embeddings of similar intervals to be closer than for different intervals. With such a model, we will have an opportunity to solve a wide range of problems. For example,
calculate similarity via various distances. 
SSL is a central approach to building informative embedding. In our case, we use non-contrastive SSL methods.





\subsection{Non-contrastive SSL}

Non-contrastive methods are based on representations of different augmentations of an object. So embeddings of different views should be close to each other in latent space. In general, such an approach can lead to a collapse of representations, i.e. algorithm will produce constant output for all inputs. Each architecture solves the collapse problem differently. Since non-contrastive approaches are based on distinct versions of each object, it is clear that augmentations are the core of such methods.

\subsection{Augmentations}
\label{sec:quality_metrics}

Augmentation strategies significantly affect the final result of SSL.
Considering the particular data, we try to select the simplest possible approach. In particular, we use jittering and window slicing as the augmentation strategies:
\begin{enumerate}
    \item Jittering or adding an independent Gaussian noise at each point is one of the simplest augmentation strategies; 
    \item Random crop is one of the most potent image augmentations. Window slicing is the term for the random crop for time series data.
\end{enumerate}

\begin{figure}
     \centering
     \includegraphics[width=0.5\textwidth]{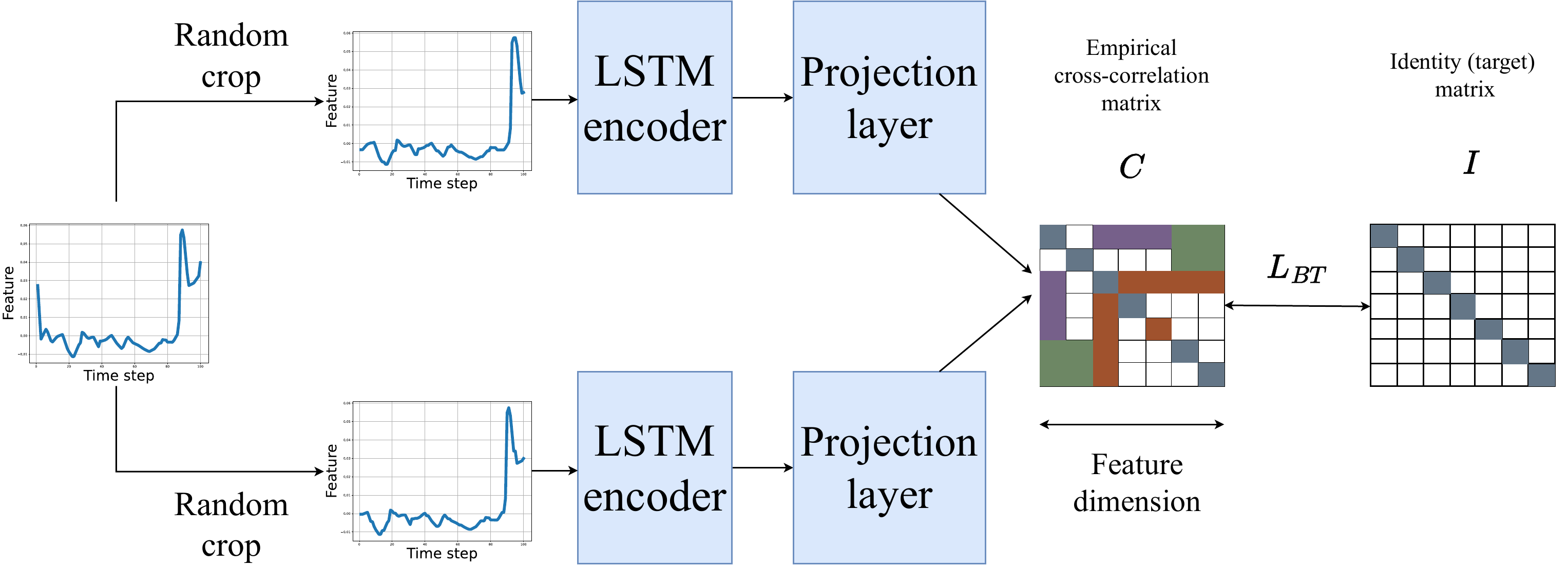}
     \caption{Self-supervised learning via Barlow Twins approach for oil\&gas data (better in zoom). The definition of $L_{BT}$ loss function is in Section ~\ref{sec:barlow twins architecture}}
     \label{fig:barlow twins architecture}
\end{figure}

\subsection{Barlow Twins}\label{sec:barlow twins architecture}

In Fig. \ref{fig:barlow twins architecture}, Barlow Twins architecture with augmentations for well-logging data can be seen. The authors of Barlow Twins~\cite{Zbontar2021} forward two views of an initial object to two neural networks with similar architecture and parameters. Both encoders are updated via gradient descent. The key idea of the method is the usage of the loss function:
$$
L_{BT} = \sum_{i} (1 - C_{ii})^2 + \lambda \sum_{i}\sum_{j \neq i} C_{ij}^2
$$
where $\lambda$ is a positive constant that balances loss terms and $C$ - cross-correlation matrix of size $d \times d$ of representation outputs of both networks over a batch, where $d$ is the embedding dimension.

The first term of the loss function enforces the closeness of embeddings of input augmentations. The second one allows decorrelation of the components of representations. 



\subsection{BYOL}
\label{sec:byol architecture}

 The BYOL method~\cite{Grill2020} has a similar architecture to Barlow Twins, but the first network called \textit{student}, has an additional predictor layer compared to the second network - \textit{teacher}. Also, 
 the weights of the teacher network are the exponential moving average (EMA) of the student network weights, so we don't update the parameters for it in the usual way. Outputs of both networks, $q$ and $z$, are \textit{$l_{2}$}-normalised: $\overline{q} = \frac{q}{{\|q\|}_2}$ and  $\overline{z} = \frac{z}{{\|z\|}_2}$. The loss function is ${\| \overline{q} - \overline{z}\|}_2^2$. To make the loss function symmetric different views are separately fed to student and teacher networks. The combination of the additional predictor in the student network and the idea of the moving average prevents collapse~\cite{Grill2020}.

\subsection{Technical details}

We use hyperparameters and procedures according to the best practices in the literature. 
The details of them are given below.

\paragraph{Encoder architecture}
More and more tasks for time series data are solved by Recurrent Neural Network (RNN)~\cite{Rumelhart1985}. RNNs often work better than traditional techniques for Time series data~\cite{Hewamalage2021}. Long short-term memory (LSTM) is a better variant of the basic RNN~\cite{Hochreiter1997}. We use the LSTM unit as an encoder for BYOL and Barlow Twins. The hidden size is 64.

\paragraph{Augmentations}
We use jittering and window slicing as augmentation techniques. 
We use a random uniform selection of an interval for slicing.
Also, an additional study on searching best hyperparameters was carried out. The standard deviation for jittering was calculated along one feature from an interval or equal to $0.03$, following the recommendations of~\cite{Iwana2021}. In the second augmentation technique, the window size changes from $25$ to $90$, with the step equal to $5$.
The procedure for selecting particular hyperparameters is in Result's section~\ref{sec:hyperparameters}.

\paragraph{Training protocol}
The duration of the training process for both architectures was the following. The maximum number of epochs was 100, like in~\cite{Chen2020}. The learning process stops if the loss function on the validation set isn't decreasing for 10 consequent epochs. To train BYOL and Barlow Twins, we mostly follow the optimization protocols from the corresponding papers. The learning rate scheduler, common for both architectures, is a cosine annealing learning rate with a maximum number of iterations $10$ and an initial learning rate $0.1$. Following ~\cite{Grill2020}, projection and prediction layers in BYOL are the same. We use $\lambda = 5 \cdot 10^{-3}$ for Barlow Twins loss function, like in~\cite{Zbontar2021}. Hidden and output dimensions of these layers for both architectures with other details are in Table \ref{tab:optimization}. We also provide the average time of one epoch that was measured for a single GeForce GTX $1080$ Ti GPU.

\begin{table}[htb]
    \centering
    \resizebox{0.5\textwidth}{!}{\begin{tabular}{l|ccc|ccc}
    \hline
      & \multicolumn{3}{c}{Architecture} & \multicolumn{3}{|c}{Optimization} \\
     \hline
     Models & hidden & output & EMA  & Batch & Optimizer & Average time (s)\\
     & dimension & dimension &  momentum & size & & one epoch\\
     \hline
     BYOL & 4096 & 256 & 0.99 & 64 &  LARS~\cite{You2017} & 16 \\
     Barlow Twins & 2048 & 2048 & absent & 2048 & same & 12 \\
     \hline
    \end{tabular}}
    \caption{Selected model characteristics}
    \label{tab:optimization}
\end{table}

\section{Data}



\subsection{Data overview}
We use an open well-logging dataset from the Taranaki Basin of New Zealand. It is provided by the New Zealand Petroleum \& Minerals Online Exploration Database~\cite{IBM2015}, and the Petlab~\cite{Strong2016}. We use $4$ different features from about $400$ wells. They are standard logs for most geophysical companies that make transfer learning easy for wells of other formations. The used 4 features are selected by experts, as others have a lot of missing values or need to be of more quality. The description of them is provided in Table~\ref{tab:features}.

\begin{table}[htb]
    \centering
    \begin{tabular}{lc}
     \hline
     Feature name & Description\\
     \hline
     DRHO & Porosity inferred from density \\
     DENS & Density \\
     GR & Gamma-ray \\
     DTC & Sonic \\
     \hline
    \end{tabular}
    \caption{Used features}
    \label{tab:features}
\end{table}

We consider intervals from different wells and their similarity. 
Each interval mathematically represents as a matrix with dimension $\mathbb{R}^{l\times d}$, where
\begin{itemize}
    \item \textit{l} - interval length;
    \item \textit{d} - number of features collected at each step of an interval.
\end{itemize}

We take $l = 100$ and $d = 4$ similar to~~\cite{Romanenkova2022}.

\subsection{Data preprocessing}

Logging data from oil \& gas wells have a complex structure with missed data, high uncertainty and other peculiarities~\cite{Acock2005}. 
So, to construct a model on top of them, we require careful preprocessing. It includes three consecutive steps: (a) filling missing values, (b) correcting sensor errors, and (c) normalization. We follow standard procedure for the data at hand~\cite{Romanenkova2022}. In particular:


\paragraph{Filling missing values} The oil \& gas industry differs in the ratio of missing values. Our data shows up to $70\%$ of missing values on each feature. Since our signals are time series, we can apply one of the most known strategies in time series for filling missing values - forward and backward fill. Our strategy to fill missing values is the following: if we have an opportunity, we use \textit{forward fill}, i.e. fill each missing value with the previously existing one. Otherwise, we use \textit{backward fill}, i.e. fill with the closest future non-missing value. 

\paragraph{Correcting sensor errors} Another issue with the logging oil \& gas data is a sensor error. Therefore, log outliers may be found in the data. The second step of data preprocessing is to drop physically inadequate data. It is necessary to drop objects where the delta between CALI (calliper) and BS (bit size) is greater than $0.35$.

\paragraph{Normalization} The final data preprocessing step is normalization~\cite{Sola1997}. GR, grouped by well and formation, is normalized by subtracting the mean and dividing by standard deviation. Other features are normalized via the whole dataset.

\section{Results}

We use clustering and classification evaluations to find the quality of self-supervised methods BYOL and Barlow Twins and compare them to the quality of baseline approaches. Before deepening into concrete results, let's describe the general evaluation strategy. 

\subsection{Evaluation strategy}

\subsubsection{Clustering}

We use a combination of class and layer expert's labelling as ground truth. More information about labelling is in~\cite{Romanenkova2022}.
As the clustering algorithm, we use agglomerative clustering that can work well for a relatively high embedding dimension. 

Clustering evaluation can be conducted via several metrics. We use a standard Adjusted Rand Index (ARI)~\cite{Rand1971}, F-score and Purity, the definition of which is described in ~\cite{zafarani2014}. The minimum and maximum of all metrics ensure to be $0$ and $1$ correspondingly.


\subsubsection{Classification}

We also evaluate BYOL and Barlow Twins via several 
downstream classification tasks on top of obtained representations.
The classifier model consists of two parts:
\begin{itemize}
    \item \textit{Encoder} has a feature description of the interval on the \textbf{input} and fixed size representation of this interval in the \textbf{output}.
    \item \textit{Classifier} receives representation from the encoder on the \textbf{input} and predicts class at the \textbf{output}.
\end{itemize}
We freeze the encoder and see if a simple classifier can solve a problem on top of it.
Thus, a better classifier score would suggest that the representations are better.

We consider Binary classification for a pair of intervals and Multiclass classification for 28, 401 and 7 classes.
The following is the formulation of the \textit{Binary classification} task, which is a type of \textit{similarity learning} also. The model considers a pair of intervals as \textit{input}. The \textit{output} should be $1$ if these intervals belong to one well and $0$ otherwise.
The \textit{Multiclass classification} model for a single \textit{input} interval should predict which well the interval belongs to. There are 401 different wells. 
We also consider a smaller set of wells of size 28 selected by an expert. The \textit{Geological classification} task is to predict the geological profile class for each well. The number of classes is 7.

\subsection{Evaluation results}

In this section, we evaluate BYOL and Barlow Twins models. We selected hyperparameters using two strategies: maximizing ARI metric (ARI) or accuracy in \textit{Geological classification} task (GEO), see Section~\ref{sec:hyperparameters}. Since we trained models with two different strategies, the names of BYOL and Barlow Twins have according clarifications in brackets. Our experiments show that both approaches have the same results for BYOL. 
We compare these architectures with top existing architectures: Triplet~\cite{Schroff2015}, Variational Autoencoder (VAE)~\cite{Egorov2022}, Simple Framework for Contrastive Learning (SimCLR)~\cite{Chen2020} and Momentum Contrast (MoCoV3)~\cite{Chen2021}. All models use a similar LSTM as an encoder. We added \textit{Rank} to the main classification and clusterization metrics, which shows the model's place among others, averaged over all metrics.

Table \ref{tab:clustering} shows that different contrastive approaches have leadership positions on various metrics, but only BYOL has the second-best result on all of them. Moreover, according to \textit{Rank}, BYOL doesn't inferior to any contrastive method and shares the first place on \textit{clusterization} task.

\begin{table}[htb]
    \centering
    \begin{tabular}{lcccc}
     \hline
     Models & ARI & Purity & F-score & Rank\\
     \hline
      BYOL(ARI/GEO) & \underline{0.36 $\pm$ 0.08} & \underline{0.45 $\pm$ 0.04} & \underline{0.36 $\pm$ 0.07} & \textbf{2.00}\\
      Barlow Twins (ARI) & 0.31 $\pm$ 0.04 & 0.34 $\pm$ 0.05 & 0.28 $\pm$ 0.04 & 5.33 \\
      Barlow Twins (GEO) & 0.23 $\pm$ 0.02 & 0.34 $\pm$ 0.06 & 0.25 $\pm$ 0.04 & 7.00 \\
      Triplet & 0.33 $\pm$ 0.03 & \textbf{0.51 $\pm$ 0.08} & \textbf{0.45 $\pm$ 0.01} & \textbf{2.00}\\
      VAE & 0.30 $\pm$ 0.03 & 0.37 $\pm$ 0.04 & 0.27 $\pm$ 0.04 & 5.67\\
      SimCLR & \textbf{0.39 $\pm$ 0.04} & 0.43 $\pm$ 0.06 & 0.34 $\pm$ 0.06 & \underline{}{2.33}\\
      MoCoV3 & 0.35 $\pm$ 0.04 &  0.42 $\pm$ 0.05 & 0.33 $\pm$ 0.06 & 3.67\\
     \hline
    \end{tabular}
    \caption{Comparison between models via clustering. Top-2 values are highlighted (the first is bold, and the second is underlined).}
    \label{tab:clustering}
\end{table}

Additional study is devoted to the examination of the classification results. 
For each classification task, we train different classifiers on top of the frozen embeddings: a logistic regression (Log.Reg.), a neural network with three fully connected networks (FC-3) and Gradient boosting (XGBoost).

The results are in Table \ref{tab:classification}. In \textit{Binary} and \textit{Geological} classification tasks BYOL and Barlow Twins (GEO) show the best performance. In other classification problems, there is no sole winner. Also, we counted \textit{Rank} for all models and tasks with XGBClassifier. The top one by Rank is Barlow Twins (GEO).


We note that our models beat alternatives on almost all tasks. As a general recommendation, based on our experiments, we can advise using a combination of XGBoost classifier on top of representations from Barlow Twins (GEO).


\begin{table*}[htb]
    \centering
    \resizebox{\textwidth}{!}{\begin{tabular}{l|ccc|ccc}
    \hline
      & \multicolumn{3}{c}{Binary classification} & \multicolumn{3}{|c}{Geological classification (7 classes)}\\
     \hline
     Models &  Log.Reg. & FC-3 & XGBoost &  Log.Reg. & FC-3 & XGBoost\\
     \hline
     BYOL(ARI/GEO) & \textbf{0.56} & \textbf{0.74} & \textbf{0.69} & 0.49 & 0.57 & 0.57\\
     Barlow Twins (ARI) & 0.53 & 0.64 & 0.63 & 0.44 & 0.48 & \underline{0.6}\\
     Barlow Twins (GEO) & \textbf{0.56} & 0.67 & \underline{0.68} & \textbf{0.62} & \textbf{0.63} & \textbf{0.82}\\
     Triplet & \underline{0.55} & 0.69 & \underline{0.68} & \underline{0.59} & \underline{0.61} & 0.59\\
     VAE & 0.54 & 0.69 & \underline{0.68} & 0.45 & 0.54 & 0.55\\
     SimCLR & 0.53 & \underline{0.73} & 0.67 & 0.50 & 0.60 & 0.56\\
     MoCoV3 & 0.53 & 0.72 & \underline{0.68} & 0.49 & 0.57 & 0.56\\
     \hline
    \end{tabular}}
    \caption{Accuracy of different models for the classification problems}
    \label{tab:classification}
\end{table*}

\subsection{Hyperparameters selection}
\label{sec:hyperparameters}

BYOL needs two distributions of image augmentations. We use window slicing as one augmentation technique and jittering as another. Unlike BYOL, the Barlow Twins approach needs two augmentations from one class. 
Since BYOL has the best results only with jittering, where the deviation is calculated via batch, and Barlow Twins works well only with window slicing, then in Table \ref{tab:hyperparameters}, we specify only window size. Chosen augmentation strategies are the best according to the simplicity-result ratio. 

\begin{table}[htb]
    \centering
    \begin{tabular}{lcc}
     \hline
     Evaluation & BYOL & Barlow Twins\\
     \hline
      Clustering (ARI) & 85 & 50 \\
      Geological classification & 85 & 65\\
     \hline
    \end{tabular}
    \caption{Selected best window size for slicing}
    \label{tab:hyperparameters}
\end{table}

\section{Conclusion}

We tackled the problem of representation learning for logging data from oil-well intervals. To address this task, we adapted self-supervised techniques from computer vision, specifically BYOL and Barlow Twins. To the best of our knowledge, our work is the first these methods were applied in the oil \& gas industry.

Special studies have been devoted to identifying optimal augmentations, which are critical components of our models and provide a simple-to-adopt solution. We also selected hyperparameters to achieve peak performance.

Our experimental results suggest the following conclusions. For the \textit{clusterization task}, BYOL has the highest Rank, and for \textit{binary classification}, BYOL outperforms all other classifiers. Notably, both problems pertain to \textit{similarity learning}. Barlow Twins embeddings produce the best results for most other \textit{classification} tasks. Furthermore, when combined with a gradient boosting classifier, this representation performs significantly better than modern baseline methods across almost all \textit{downstream} tasks.


So, our non-contrastive approaches and the right augmentation strategy provide better representations than other self-supervised methods. Thus, they are universal and suitable for the solution of diverse problems.


\bibliography{mybibfile}

\end{document}